\documentclass[letterpaper, 10 pt, conference]{ieeeconf}  

\IEEEoverridecommandlockouts                              

\usepackage{amsmath} 
\usepackage{amssymb}  

\usepackage{amsthm}
\theoremstyle{definition}
\newtheorem{definition}{Definition}[section]

\usepackage{xcolor}
\definecolor{wong-black}        {HTML}{000000}
\definecolor{wong-lightorange}  {HTML}{E69F00}
\definecolor{wong-lightblue}    {HTML}{56B4E9}
\definecolor{wong-green}        {HTML}{009E73}
\definecolor{wong-yellow}       {HTML}{F0E442}
\definecolor{wong-darkblue}     {HTML}{0072B2}
\definecolor{wong-darkorange}   {HTML}{D55E00}
\definecolor{wong-pink}         {HTML}{CC79A7}
\usepackage{hyperref}
\usepackage{tikz}
\usepackage{etoolbox}
\usepackage{color}
\usetikzlibrary{shapes.geometric, arrows, positioning}
\usepackage{caption}
\usepackage{subcaption}
\usepackage{graphicx}
\usepackage{booktabs}
\usepackage{url}

\title{\LARGE \bf
Informed Reinforcement Learning for\\Situation-Aware Traffic Rule Exceptions
}

\author{Daniel Bogdoll$^{1,2,*}$, Jing Qin$^{2,*}$, Moritz Nekolla$^{1}$,\\Ahmed Abouelazm$^{1}$, Tim Joseph$^{1}$, and J. Marius Zöllner$^{1,2}$
\thanks{$^{1}$Authors are with the FZI Research Center for Information Technology, Germany
        {\tt\small bogdoll@fzi.de}}%
\thanks{$^{2}$Authors are with the Karlsruhe Institute of Technology, Germany}%
\thanks{$^{*}$These authors contributed equally}%
}

\begin{document}

\maketitle
\thispagestyle{empty}
\pagestyle{empty}

\begin{abstract}

    Reinforcement Learning is a highly active re-
search field with promising advancements. In the field of
autonomous driving, however, often very simple scenarios are
being examined. Common approaches use non-interpretable
control commands as the action space and unstructured reward
designs, which are unsuitable for complex scenarios. In this work, we introduce
Informed Reinforcement Learning, where a structured rulebook
is integrated as a knowledge source. We learn trajectories and
asses them with a situation-aware reward design, leading to
a dynamic reward that allows the agent to learn situations
that require controlled traffic rule exceptions. Our method is
applicable to arbitrary RL models. We successfully demonstrate
high completion rates of complex scenarios with recent model-based agents.

\end{abstract}

\section{Introduction}
\label{sec:introduction}

The rapid development of autonomous driving has led to multiple SAE Level 4 fleets available to the public in small, restricted Operational Design Domains~\cite{sf_level4}. The high flexibility necessary to progress in complex traffic scenarios is especially challenging~\cite{waymax}. In research, often simple scenarios are being examined, which can hardly be transferred to real-world scenarios~\cite{Bogdoll_Ontology_2022_ECCV}. Especially hierarchical traffic rules, which sometimes override others in specific situations, are often neglected but are a typical occurrence in everyday traffic~\cite{censiLiabilityEthicsCultureAware2019}. Especially in the field of Reinforcement Learning, which has shown rapid advancements, often very simple and conflicting reward functions are being used in the domain of autonomous driving, which do not have the potential to solve such challenges~\cite{KNOX2023103829}.

\textbf{Research Gap.}
While the development of autonomous vehicles has rapidly progressed, their ability to comprehend hierarchical traffic rules, i.e., rules that override others in certain situations, remains a challenge~\cite{censiLiabilityEthicsCultureAware2019}. Current methods predominantly focus on standard traffic scenarios~\cite{aradiSurveyDeepReinforcement2022}, often neglecting the nuances of exception handling. Additionally, even though Reinforcement Learning~(RL) has made strides in behavior planning and control instructions for autonomous driving, the potential of RL in direct trajectory generation is not extensively researched~\cite{moghadamEndtoendDeepReinforcement2020}.

\textbf{Contribution.} Our work leverages the capabilities of Informed Reinforcement Learning~\cite{wormannKnowledgeAugmentedMachine2023} to enhance the decision-making and adaptability of autonomous vehicles, especially in traffic rule exception scenarios. The core contributions are:

\begin{itemize}
    \item Introduction of a \textit{rulebook} in RL reward design: A structured, machine-comprehensible encoding of hierarchical traffic rules, leading to a situation-aware reward
    \item Our work is the first one to successfully learn \textit{feasible trajectories} in Frenet space purely based on raw sensory RGB observations of the environment
    \item We provide a benchmark of 1,000 \textit{anomaly scenarios} in the CARLA simulation environment, where the agent must be able to execute controlled rule exceptions to reach the goal
\end{itemize}

More details are available in~\cite{Qin_Reinforcement_2023_MA}, all code is on \href{https://github.com/fzi-forschungszentrum-informatik/informed_rl}{GitHub}.

\tikzstyle{module} = [rectangle, rounded corners, minimum width=2cm, minimum height=0.7cm,text centered, text width=2cm, draw=black]
\tikzstyle{data} = [cylinder, minimum width=0cm, minimum height=0cm, text centered, draw=black, fill=white, text width=2cm, shape border rotate=90, aspect=0.1]

\begin{figure}[t!]
    \begin{center}
        \begin{tikzpicture}[node distance=1.3cm, auto]
            \node [draw, data, fill=wong-green!20] (anomaly) {Anomaly\\Scenarios};
            \node [draw, data, above of=anomaly,yshift=+0.5cm] (normal) {Normal\\Scenarios};
            \node [draw, module, right of=anomaly,xshift=2cm, fill=wong-green!20] (dreamer) {RL Agent};
            \node [draw, module, above of=dreamer, yshift=+0.5cm] (rulebook) {Hierarchical\\Rulebook};
            \node [draw, module, right of=dreamer,xshift=2cm, fill=wong-green!20] (pid) {Controller};
            \node [draw, module, above of=pid, yshift=+0.5cm, fill=wong-green!20] (traj) {Trajectory Generation};

            \draw[->] (traj.west) -- node[swap] {$\tau_t$} (rulebook.east); 
            \draw[->] (anomaly) -- node {$o_t$} (dreamer);
            \draw[->] (normal) -- node {$\hat{o}_t$} (rulebook);
            \draw[-] (anomaly.east) -- + (0.25,0)
            |- node[pos=0.3, right] {} (rulebook.west);
            \draw[->] (traj) -- node[] {$\tau_t$} (pid);
            \draw[->] (rulebook) -- node[] {$r_{RB,t}$} (dreamer);


            \draw[->] (dreamer.east) -- + (0.25,0)
            |- node[pos=0.26, right] {$a_t$} ([yshift=-0.2cm]traj.west);

            \draw[-] (normal.east) -- + (0.25,0)
            |- node[pos=0.3, right] {} (dreamer.west);
        \end{tikzpicture}
    \end{center}
    \caption{Architecture of our approach. We use Curriculum Learning, where normal scenarios are used first to learn basic driving behavior. Then, anomalies are provided to learn controlled rule exceptions. Given an observation $o_t$, the Reinforcement Learning agent chooses an action $a_t$ as the parametric input for generating a trajectory $\tau_t$. The rulebook then evaluates the trajectory in the context of an abstracted environment $\hat{o}_t$ and provides the partial reward $r_{RB,t}$. Finally, a controller follows the trajectory. During evaluation, only the path in green is executed.}
    \label{fig:pipeline}
\end{figure}
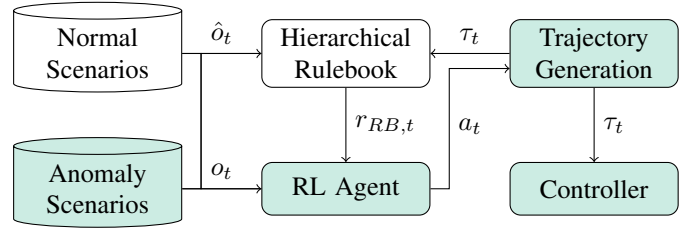
\section{Related Work}
\label{sec:related_work}
In this section, we review related work on the application of Reinforcement Learning (RL) in motion planning for self-driving vehicles, typical traffic scenarios for training, and traffic rule formalization.

\subsection{Reinforcement Learning for Motion Planning}
Motion planning in the context of autonomous driving can be split into behavioral planning, trajectory planning, and control instructions~\cite{aradiSurveyDeepReinforcement2022}.

In the context of \textit{behavioral planning}, Fayjie et al.~\cite{fayjieDriverlessCarAutonomous2018} proposed an RL-based autonomous driving strategy for urban traffic scenarios, where the discrete action space consists of \textit{left}, \textit{right}, and \textit{keep going} to symbolize lane changing behaviors. Ye et al.~\cite{yeAutomatedLaneChange2020} proposed a strategy for automatic lane changing with RL based on Proximal Policy Optimization. This strategy enables a trained agent to make efficient lane-changing decisions even in dense traffic scenarios. Furthermore, numerous studies have been conducted exploring RL-based behavioral planning for autonomous vehicles, with findings demonstrating reliable performance~\cite{huInteractionawareDecisionMaking2020, boutonReinforcementLearningIterative2020, caoHighwayExitingPlanner2021, hoelAutomatedSpeedLane2018, guoMergingDivergingImpact2021}.

In the context of \textit{trajectory planning}, Feher et al.~\cite{feherHybridDDPGApproach2019} learned waypoints an agent should follow. For that, they used the Deep Deterministic Policy Gradient (DDPG) algorithm. A limitation of this methodology lies in its sole focus on lateral planning. Moghadam et al.~\cite{moghadamEndtoendDeepReinforcement2020} proposed an RL agent that learns input parameters for a trajectory planner on the Frenet Space for highway scenarios. They use a continuous action space with processed time-series data as observation space instead of raw sensory observations. Coad et al.~\cite{coad2020safe} presented a continuous RL agent in a static occupancy grid. The agent's action space is a sequence of changes in curvilinear coordinates, lateral displacement, and velocity with a fixed longitudinal step. Lu et al.~\cite{lu2023action} proposed a hierarchical reinforcement learning framework for trajectory planning given a state space composed of BEV images and lidar data. The framework consists of a high-level action responsible for choosing the direction of motion and a medium-level action sampling the vehicle's next waypoint from a fixed-size semi-circle, which can also sample off-road waypoints.

In the aerospace sector, Goddard et al.~\cite{goddardUtilizingReinforcementLearning2022} proposed a framework that adds recorded flight trajectories from and after training to a database of motion primitives, which are then processed and used for subsequent classical motion planning. Williams et al.~\cite{williamsTrajectoryPlanningDeep2022} propose a similar method for space-shuttle trajectories, where their RL agent learns the parameters of motion primitives, i.e., the length and defining node points of splines, which are then combined into a full trajectory. Only during training does the agent learn to avoid infeasible trajectories, as the action space does not guarantee this.

In the context of \textit{control instructions}, where we focus on end-to-end learning, many methods leverage raw sensor inputs as the input space and directly output control commands for autonomous vehicle control, which include steering angle and acceleration~\cite{aradiPolicyGradientBased2018, jaritzEndtoEndRaceDriving2018, nageshraoAutonomousHighwayDriving2019, sallabEndtoEndDeepReinforcement2016, yuIntelligentLandVehicleModel2018}. However, these approaches are challenging to interpret, which makes them difficult to apply in real-world scenarios.

\subsection{Traffic Scenarios}
Most RL-based autonomous driving studies set up a specific autonomous driving environment for the vehicle. Given the relatively straightforward nature of \textit{highway traffic} conditions, these environments present comparatively less complex scenarios for autonomous driving. Consequently, A substantial number of studies has opted to utilize highway scenarios as the benchmark for evaluating RL-based autonomous driving strategies~\cite{xuReinforcementLearningApproach2020, aradiPolicyGradientBased2018, baiDeepReinforcementLearning2019a, hoelAutomatedSpeedLane2018, nageshraoAutonomousHighwayDriving2019, wolfAdaptiveBehaviorGeneration2018}.

However, further approaches have focused on \textit{urban area traffic}, encompassing elementary urban traffic, intersections, as well as dense urban traffic situations~\cite{boutonReinforcementLearningIterative2020, fayjieDriverlessCarAutonomous2018, yeAutomatedLaneChange2020, zhangSafeRuleAwareDeep2022}.

Nonetheless, there is a lack of literature considering traffic rule exception scenarios~\cite{aradiSurveyDeepReinforcement2022,Bogdoll_Ontology_2022_ECCV}. Talamini et al.~\cite{talaminiImpactRulesAutonomous2020} utilize RL to train a driving strategy when controlled traffic rule exceptions become necessary. Their approach considers behavior planning with lateral motion only and does not provide a structured approach regarding the integration of rules into the reward.

\subsection{Formalism of Traffic Rules}
Many studies have explored formalizing traffic rules into a machine-readable format. A number of methods has been used, e.g., temporal logic~\cite{maierhoferFormalizationInterstateTraffic2020}, linear temporal logic (LTL)~\cite{kloetzerFullyAutomatedFramework2008}, signal temporal logic (STL)~\cite{aguilarSTLRulebooksRewards2021}, Isabelle theorem proving~\cite{rizaldiFormalisingMonitoringTraffic}, and fuzzy logic~\cite{morseFuzzyApproachQualification2017}. However, these studies primarily concentrate on translating individual rules without considering the prioritization among different rules. Censi et al.~\cite{censiLiabilityEthicsCultureAware2019} introduced a theoretical “rulebook” to structure different rules, establish a hierarchy between rules, and analyze traffic rule exception scenarios, but did not provide a framework or implementation to actually utilize it.

Consequently, there is a noticeable research gap in the development of an RL application for autonomous vehicles that not only addresses the trajectory generation in traffic rule exception scenarios, but also efficiently incorporates a structured set of traffic rules into the reward function.
\section{Method}
\label{sec:method}

This section presents our methodology as visualized in Fig.~\ref{fig:pipeline}. We first introduce the problem statement that outlines the challenges discussed in this paper. Next, the generation of vehicle trajectories using the Frenet Space~\cite{frenet_space} is detailed. Subsequently, the process of structuring traffic rules for computational interpretation is discussed, including using the rulebook and its integration into a reward function. Our methodology can be utilized with arbitrary RL agents.

\subsection{Problem Statement}
We focus on handling scenarios requiring controlled traffic rule exceptions, as discussed in Section~\ref{sec:introduction}. These situations contain apparent conflicts of traffic rules, where one rule can override another. For an agent to solve the tasks, situation-awareness is necessary to apply the correct set of rules at any given time. As the system dynamics are unknown, we model the problem as a Partially Observable Markov Decision Process (POMDP), with the state space comprising high-dimensional sensory RGB data from a BEV camera and the action space consisting input parameters for the generation of a trajectory in Frenet Space.

\subsection{Trajectory Generation}
Following the approach presented by Werling et al.~\cite{werlingOptimalTrajectoriesTimecritical2012}, we generate trajectories in Frenet Space. For this, terminal manifolds, symbolized as $A\{v, d, t\}$, are altered according to the current state of the vehicle:

\begin{itemize}
    \item $v$: Desired velocity at the termination of the trajectory
    \item $d$: Lateral offset relative to the reference trajectory
    \item $t$: Time needed to reach the desired target state
\end{itemize}

The goal state of a trajectory in Frenet space is fully defined by $v, d, t$, which means that each set of parameters corresponds to a trajectory. As shown in Fig.~\ref{fig:pipeline}, a controller is then utilized to follow the generated trajectory.

\subsection{Situation-Aware Reward Design}
In the case of traffic rule exception scenarios, some rules can override others. For this, we present a formal rulebook as part of the reward function to represent situation-specific hierarchies between different rules. Assuming no rule conflicts as the default state, the agent first needs to assess the current situation to activate dynamic rewards.

\textbf{Situation Awareness.}
In order to assess traffic situations, an agent needs to have an understanding of traffic rules and capabilities to monitor them~\cite{Bogdoll_Quantification_2022_IV}. As shown in Fig.~\ref{fig:pipeline}, the abstracted environment $\hat{o}_t$ provides information for this purpose, such as map data. Similar to Censi et al.~\cite{censiLiabilityEthicsCultureAware2019}, we introduce \textit{Rule Realizations}:

\begin{definition}[Rule Realization]
    Let $\tau_t$ be a sequence of states, i.e., a trajectory. A rule realization $\psi: \tau_t \rightarrow \mathbb{R}$ assigns a real number $\psi(\tau_t)$ to $\tau_t$ . This is an expression of the degree of compliance of $\tau_t$ with an underlying traffic rule.
\end{definition}

Based on this knowledge of existing traffic rules, an agent can then set rule coefficients $\rho_j$ depending on the current situation, e.g., diminishing the relevance of a rule if it is overwritten by another one.


\textbf{Hierarchical Rulebook.}
Inspired by the conceptual hierarchical rulebook by Censi et al.~\cite{censiLiabilityEthicsCultureAware2019}, we present an implementation of a rulebook within the reward function of an RL agent. We define a rulebook as follows:

\begin{definition}[Rulebook]
    A rulebook $\mathcal{R}$ is defined as a tuple $(\Psi, \preceq)$, where $\Psi$ represents a finite set of rule realizations and $\preceq$ denotes a pre-order relation. Specifically, $\psi \preceq \psi'$ implies that $ \psi$ is of a lower hierarchy than $\psi'$.
\end{definition}

We utilize Linear Temporal Logic~(LTL) syntax for rule descriptions and their integration into the reward function. The rulebook is only activated when the situation awareness module detects a situation where a controlled rule exception becomes possible. The hierarchical structure of the rulebook is instrumental in determining which rules have precedence over others. Its hierarchical structure can be visualized as a graph, with each rule realization as a node and edges indicating priority relationships. Nodes of equal priority can be merged.

\tikzstyle{rule} = [rectangle, rounded corners, minimum width=1.5cm, minimum height=0.7cm,text centered, text width=1cm, draw=black]

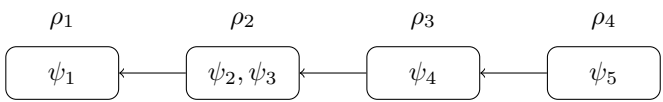
\begin{figure}[h]
    \begin{center}
        \begin{tikzpicture}[node distance=1.4cm, auto]
            \node [draw, rule] (one) {$\psi_1$};
            \node [draw, rule, right of=one, xshift=1cm] (two_three) {$\psi_2,\psi_3$};
            \node [draw, rule, right of=two_three, xshift=1cm] (four) {$\psi_4$};
            \node [draw, rule, right of=four, xshift=1cm] (five) {$\psi_5$};

            \node [rectangle, above of=one, yshift=-0.7cm] (one_p) {$\rho_1$};
            \node [rectangle, above of=two_three, yshift=-0.7cm] (two_p) {$\rho_2$};
            \node [rectangle, above of=four, yshift=-0.7cm] (three_p) {$\rho_3$};
            \node [rectangle, above of=five, yshift=-0.7cm] (four_p) {$\rho_4$};

            \draw[->] (five) -- node {} (four);
            \draw[->] (four) -- node {} (two_three);
            \draw[->] (two_three) -- node {} (one);
        \end{tikzpicture}
    \end{center}
    \caption{Graph representation of a hierarchical rulebook $\Psi$ with rule realizations $\psi_i$ and hierarchy coefficients $\rho_j$, where $j$ indicates the hierarchy index.}
    \label{fig:rulebook_coeff}
\end{figure}

For instance, in Fig.~\ref{fig:rulebook_coeff}, $\psi_1$ holds the highest hierarchy, $\psi_2$ and $\psi_3$ share the same, followed by $\psi_4$, and $\psi_5$ with the lowest. Such a representation assures the rulebook's scalability.

\textbf{Linear Temporal Logic.}
LTL is a powerful logic language utilized in defining sequences of events or states. Its syntax contains various logical operators: negation ($\neg$), conjunction ($\land$), disjunction ($\lor$), and implication ($\rightarrow$), along with temporal operators: \textit{Next} ($\mathbf{X}$), \textit{Globally} ($\mathbf{G}$), \textit{Finally} ($\mathbf{F}$), and \textit{Until} ($\mathbf{U}$). To evaluate a trajectory $\tau_t$, which can be defined as a sequence of vehicle states, we apply LTL for each rule realization $\psi_i$. This reward calculation can be described by a function $f(\psi_i,\tau_t)$.

\textbf{Reward Design.}
The rulebook's hierarchical structure is incorporated into the reward function based on hierarchy coefficients $\rho_j \in [0,1]$ for each hierarchy, as shown in Fig.~\ref{fig:rulebook_coeff}. The coefficient scales the reward or penalty associated with a given level's rules so that higher-hierarchy rules have a higher weight in the reward. This way, the reward is dynamically adapted to the current situation. All coefficients default to 1 in standard scenarios. The realization is shown in Eq.~\eqref{eq:rulebook_reward}.

\begin{equation}    r_{RB,t}=\sum_{\psi_i\in\Psi}\left(\prod_{\psi_j=\psi_i}^{\psi^{\prime}}\rho_j\right)\cdot f(\psi_i,\tau_t)\cdot c_{\psi i}
    \label{eq:rulebook_reward}
\end{equation}

Here, $\Psi$ are all the rules involved in the rule exception, $\psi^{\prime}$ is the rule with the highest priority in the rulebook, and $\rho_j$ is the hierarchy coefficient of $\psi_j$. $f(\psi_i,\tau_t)$ is the reward value. $c_{\psi i}$ is a scaling factor for each rule that allows fine-tuning.
\section{Experiments}
\label{sec:experiments}
This section presents our experimental setup for the training and evaluation of our Reinforcement Learning agents. We present the traffic scenarios and rules we examined, the state and action spaces, our reward function, as well as the training process and parameters.

\begin{figure}[h]
    \centering
    \includegraphics[width=\columnwidth]{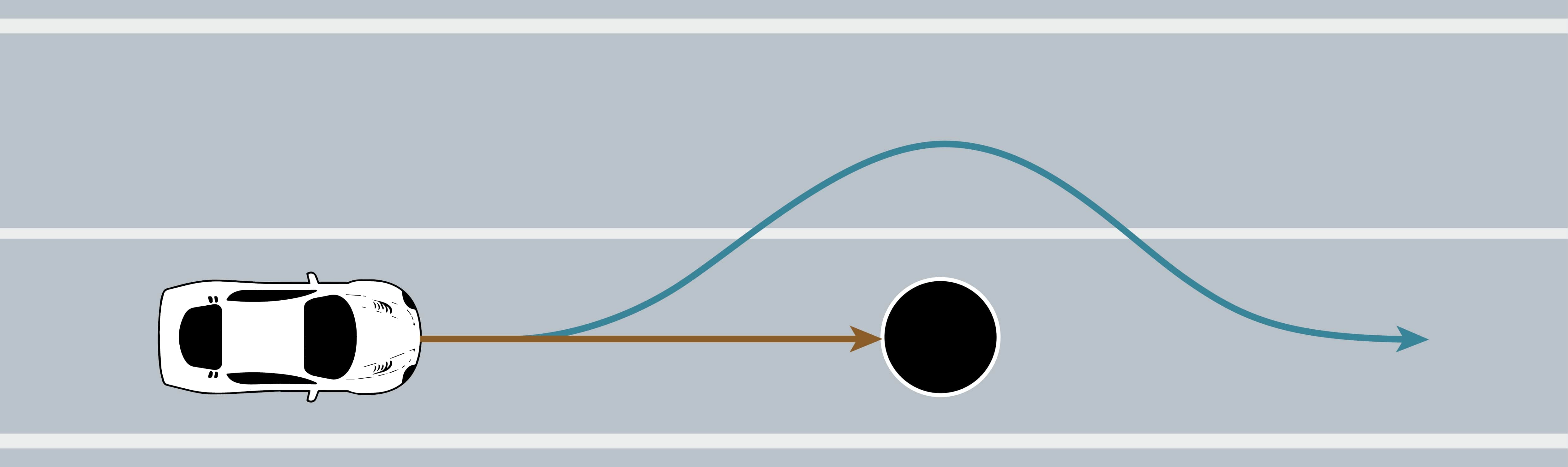}
    \caption{Traffic scenario that shows an atypical scenario with an anomaly. In the illustrated scenario, the ego vehicle's lane is blocked, enabling it to perform a controlled rule exception. Adapted from~\cite{Qin_Reinforcement_2023_MA}.}
    \label{fig:rule_excep}
\end{figure}

\textbf{Traffic Scenarios.} For our experiments, we chose a typical scenario in everyday urban traffic. Based on the German road traffic regulations, it is generally forbidden to cross a solid line. However, in certain situations, e.g., when the lane of the ego vehicle is blocked, there exists a rule exception~\cite{durchgezogenelinie}, as shown in Fig.~\ref{fig:rule_excep}.

We provide a benchmark with 1,000 such scenarios in the CARLA simulation environment~\cite{Dosovitskiy17} and also the codebase to generate more if needed. Each scenario is defined by a reference trajectory with a length of 80 meters and an arbitrary object that blocks the lane at some point along the trajectory, as shown exemplary in~Fig~\ref{fig:DreamerV3_prediction}.

\textbf{Agent Selection.} For our experiments, we selected two established Reinforcement Learning Models. First, we decided to utilize the current state-of-the-art model-based algorithm DreamerV3, which demonstrated superior performance in a wide variety of domains~\cite{hafnerMasteringDiverseDomains2023}. Second, we utilized the model-free Rainbow algorithm~\cite{hesselRainbowCombiningImprovements2017}, an improved version of the well-known Deep Q-Networks (DQN). In both cases, CNNs were used to encode the observations.

\textbf{State Space.} The state space comprises BEV RGB images with a resolution of $128x128$, as shown in~Fig.~\ref{fig:DreamerV3_prediction}. This state space inherently captures essential aspects like road geometry, the ego vehicle's position, obstacles, and the planned path.

\begin{figure}[t!]
    \centering
    \begin{subfigure}[t]{0.5\columnwidth}
        \centering
        \includegraphics[width=0.95\columnwidth]{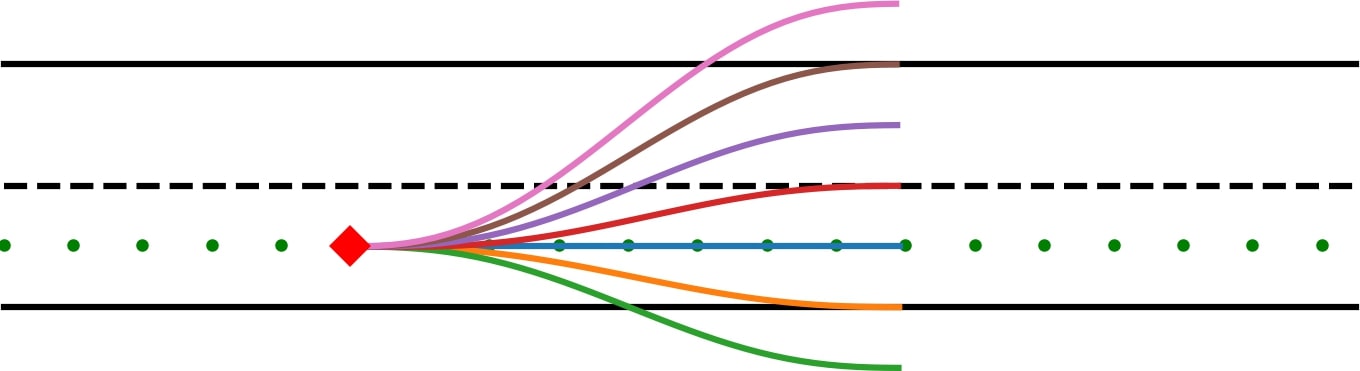}
        \caption{Action space for $d=0$}
    \end{subfigure}%
    \begin{subfigure}[t]{0.5\columnwidth}
        \centering
        \includegraphics[width=0.95\columnwidth]{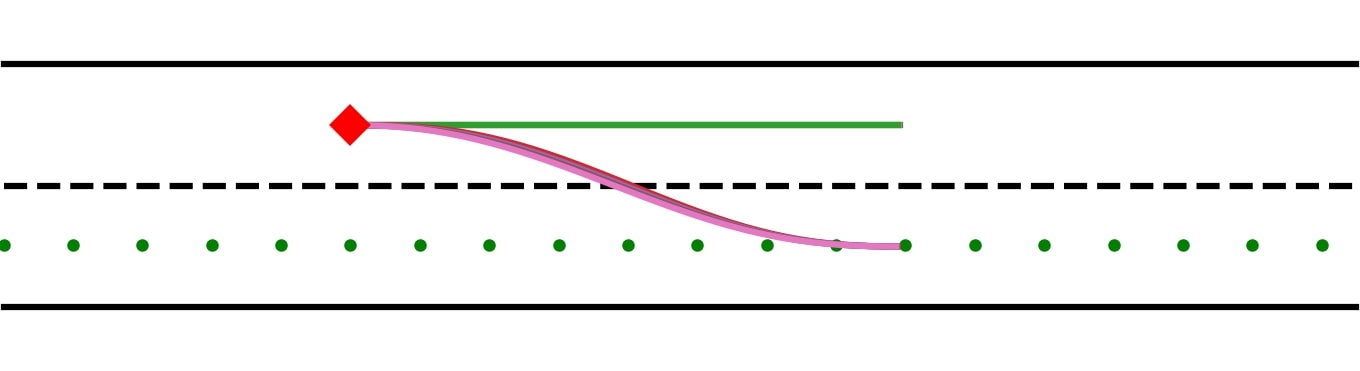}
        \caption{Action space for $d\neq 0$}
    \end{subfigure}
    \caption{Dynamic action space in Frenet space. The left side shows a scenarios where the ego vehicle is in its intended lane while it is in the opposite lane on the right side.}
    \label{fig:action_space}
\end{figure}

\textbf{Action Space.} In order to generate trajectories, we utilize the terminal manifold $A\{v, d, t\}$ introduced earlier. Here, $d$ helps to avoid obstacles, while $v$ and $t$ affect trajectory curvature and length. These translate into full trajectories in Frenet space. As we are most interested in the ability of the agent to avoid obstacles, we simplify the discrete action space. We set $v$ and $t$ to constants and $d$ to specific values in dependence on the vehicle's position, as illustrated in Fig.~\ref{fig:action_space}. As shown in Fig.~\ref{fig:pipeline}, a PID controller is implemented to follow the generated trajectory.

\textbf{Situation-Aware Reward.} For the total reward $r_t$, we combine two aspects, as shown in Eq.~\eqref{eq:reward}. The first component utilizes the current state of the ego vehicle and the second one is based on our situation-aware rulebook.

\begin{equation}
    r_t = r_{ego,t} + r_{RB,t}
    \label{eq:reward}
\end{equation}

The first component $r_{ego}$ is shown in Eq.~\eqref{eq:r_ego}. It consists of $r_{finish} = 10$ if the vehicle reaches $s_{target}$ but not $d_{target}$, $60$ if both are reached and $0$ otherwise. Additionally, $r_{speed} = -1$ if the speed is not within $10-50km/h$, otherwise, it is $0$. The trajectory length traveled in the past step is denoted by $l$. All values were determined by a small set of experiments.

\begin{equation}
    r_{ego} = r_{finish} + r_{speed} * l
    \label{eq:r_ego}
\end{equation}

For the second reward component $r_{RB,t}$, we utilize a set of simplified traffic rules necessary for the designed scenarios. As shown in Table~\ref{table:reward_rulebook}, our agents monitor three rules, focussing on collision avoidance and adherence to road layout. Per default, the agent should adhere to all rules.

\begin{table}[h!]
    \centering
    \resizebox{\columnwidth}{!}{%
        \begin{tabular}{@{}lllcr@{}}
            \toprule
            Rule             & Realization & LTL Formula      & $j$ & $\rho_j$ \\ \midrule
            Avoid collisions & $\psi_1$    & G(no\_collision) & 1   & 1        \\
            Stay in lane     & $\psi_2$    & G(in\_lane)      & 2   & 0.1      \\
            Stay on road     & $\psi_3$    & G(no\_out\_road) & 2   & 0.1      \\
        \end{tabular}%
    }
    \caption{Rule overview  including rule realization IDs, LTL formulas, hierarchy levels $j$ and coefficients $\rho_j$.}
    \label{table:reward_rulebook}
\end{table}

All rules can be monitored based on the temporal operator~$\mathbf{G}$ from LTL, as introduced in Section~\ref{sec:method}. As shown in Eq.~\eqref{eq:f1}, states breaking the rule receive a penalty of $-1$ per rule realization, otherwise $0$. The expression $\tau_t \not\models \mathbf{G}\psi$ refers to whether the states in the generated trajectory satisfy a rule. The trajectory is considered to violate a rule if any state does not satisfy it.

\begin{equation}
    f(\mathrm{G}\psi,\tau_t)=
    \begin{cases}
        -1, & \text{if } \tau_t \not\models \mathrm{G}\psi \\
        0,  & \text{Otherwise}
    \end{cases}
    \label{eq:f1}
\end{equation}

Given the concrete set of rules and rule realizations from Table~\ref{table:reward_rulebook} and the rulebook reward as defined in Eq.~\eqref{eq:rulebook_reward}, we express $r_{RB,t}$ as follows:

\begin{equation}
    \begin{aligned}
        r_{RB,t}
         & = \rho_1 * r_{collision} * c_{col}              \\
         & + \rho_1 * \rho_2 * r_{in\_lane} * l * c_{lane} \\
         & + \rho_1 * \rho_2 * r_{no\_out\_road} * l
    \end{aligned}
    \label{eq:r_rulebook}
\end{equation}

When the scenarios demand it, controlled rule exceptions become necessary to proceed. Based on ground truth, the situation awareness module of the agent activates the rulebook when it approaches an obstacle. This becomes evident in Eq~\eqref{eq:r_rulebook}, where all coeffients $\rho_j$ are then set to their values as defined in Table~\ref{table:reward_rulebook} instead of their default value $1$. Thus, when necessary, the agent can leave the road with only minor negative influence on the reward in order to perform a controlled rule exception.

\textbf{Curriculum Learning.} We divided our training strategy into two steps. First, the agent shall learn regular driving behavior. After that, we introduce situations that require controlled traffic rule exceptions, as shown in~Fig.~\ref{fig:pipeline}. Thus, for the first 3,000 steps, our agent is trained in a simple urban environment. Subsequently, we continue the training with scenarios that require the previously introduced traffic rule exceptions. The key training parameters are detailed in Table \ref{table:train_parameters}.


\begin{table}[h!]
    \centering
    \resizebox{\columnwidth}{!}{%
        \begin{tabular}{@{}llll@{}}
            \toprule
            Parameters         & Value   & Parameters    & Value         \\ \midrule
            steps              & $4e4$   & device        & A100          \\
            action repeat      & 1       & obs\_size     & {[}128,128{]} \\
            train\_scenarios        & 800     & batch\_size   & 32            \\
            actor\_dist        & one hot      & batch\_length & 64            \\
            value\_lr          & 3e-5 & discount      & 0.997         \\
            model\_lr          & 1e-4    & actor\_lr     & 3e-5          \\
            
        \end{tabular}%
    }
    \caption{Training parameters}
    \label{table:train_parameters}
\end{table}
\section{Evaluation}
\label{sec:evaluation}
In this section, we compare and analyze the results of our experiments. We compare two Reinforcement Learning Agents and perform a variety of ablation studies in order to attribute the performance of our approach to the individual components. We show both quantitative results for the whole training process as well as qualitative demonstrations of how our most successful agent performs in scenarios that require controlled traffic rule exceptions.

\subsection{Quantitative Evaluation}
For the evaluation, we opted for two key metrics based on the vehicle's performance in avoiding obstacles, returning to the original lane, and adhering to traffic rules:
\begin{itemize}
    \item \textbf{Arrived Distance:} The distance the vehicle was able to travel along the s-axis in the Frenet coordinate at the end of each episode, reflecting the distance traveled along the lane.
    \item \textbf{Finished Score:} Value ranging from 0 to 1 that quantifies the success in completing the anomaly scenario navigation task. A value of 1 denotes full success, 0.5 indicates returning to the correct longitudinal but not lateral position, and 0 is assigned otherwise.
\end{itemize}

These metrics collectively assess the agent's ability to manage scenarios which require controlled traffic rule exception.

As we extended existing RL models with our trajectory generation component and the situation-aware reward design, we performed four types of \textit{ablation studies}:

\begin{itemize}
    \item Baseline: Here we implemented the baseline RL agents in an end-to-end setting with a discrete control based action space, consisting of three acceleration values $(-1, 0, 1)$ and three angular velocity values $(-1, 0, 1)$.
    \item Trajectory: Here, we implemented only the trajectory generation without the situation-aware reward. This means that all coefficients $\rho_j$ are set to $1$ constantly.
    \item Rulebook: Here, we implemented only the situation-aware reward function but did not utilize our trajectory generation. In this case, Eq.~\eqref{eq:f1} will only check the state of the vehicle at each timestep
    \item Combination: Here, we implemented both the trajectory generation and the situation-aware reward function.
\end{itemize}

\begin{figure}[t]
    \begin{center}
        \resizebox{\columnwidth}{!}{\input{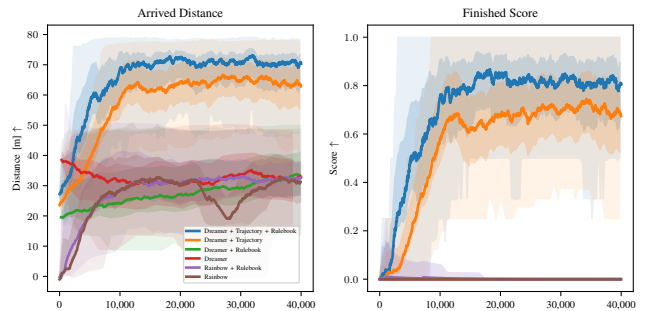}}
    \end{center}
    \caption{Evaluation of the \textit{arrived distance} and \textit{finished score} during training, showing the running average, standard deviation, and 5th and 95th percentiles. We compared agents that worked with direct controls as their output or a \textit{trajectory} and utilized either a conservative reward or our \textit{rulebook}.}
    \label{fig:Training_curve}
\end{figure}

We trained a total of six different models in accordance with our ablation study design, as shown in Fig.~\ref{fig:Training_curve}. Independant of the underlying RL model, the scenarios in combination with our reward were too challenging for both baseline models. Including only the rulebook had no clear influence. At this point we decided to focus on the generally more capable DreamerV3 agent. In combination with our trajectory planning module, DreamerV3 consistently outperformed other methods on both metrics \textbf{Arrived Distance} and \textbf{Finished Score}. This approach exhibited a steeper learning curve, suggesting rapid adaptation to guide the vehicle efficiently. When we additionally activated our situation-aware reward function, the performance of the DreamerV3 model improved further. This indicates that the reward function is beneficial for the agent's learning process, as the total performance stas consistently above the approach without the situation-awareness, while not achieving a 100\% success rate in either of the metrics. In order to better understand failure cases, we have also performed a qualitative evaluation, which will be presented in the following section.


\subsection{Qualitative Evaluation}
\begin{figure}[t!]
    \centering
    \begin{subfigure}[b]{0.49\columnwidth}
        \centering
        \includegraphics[width=\textwidth]{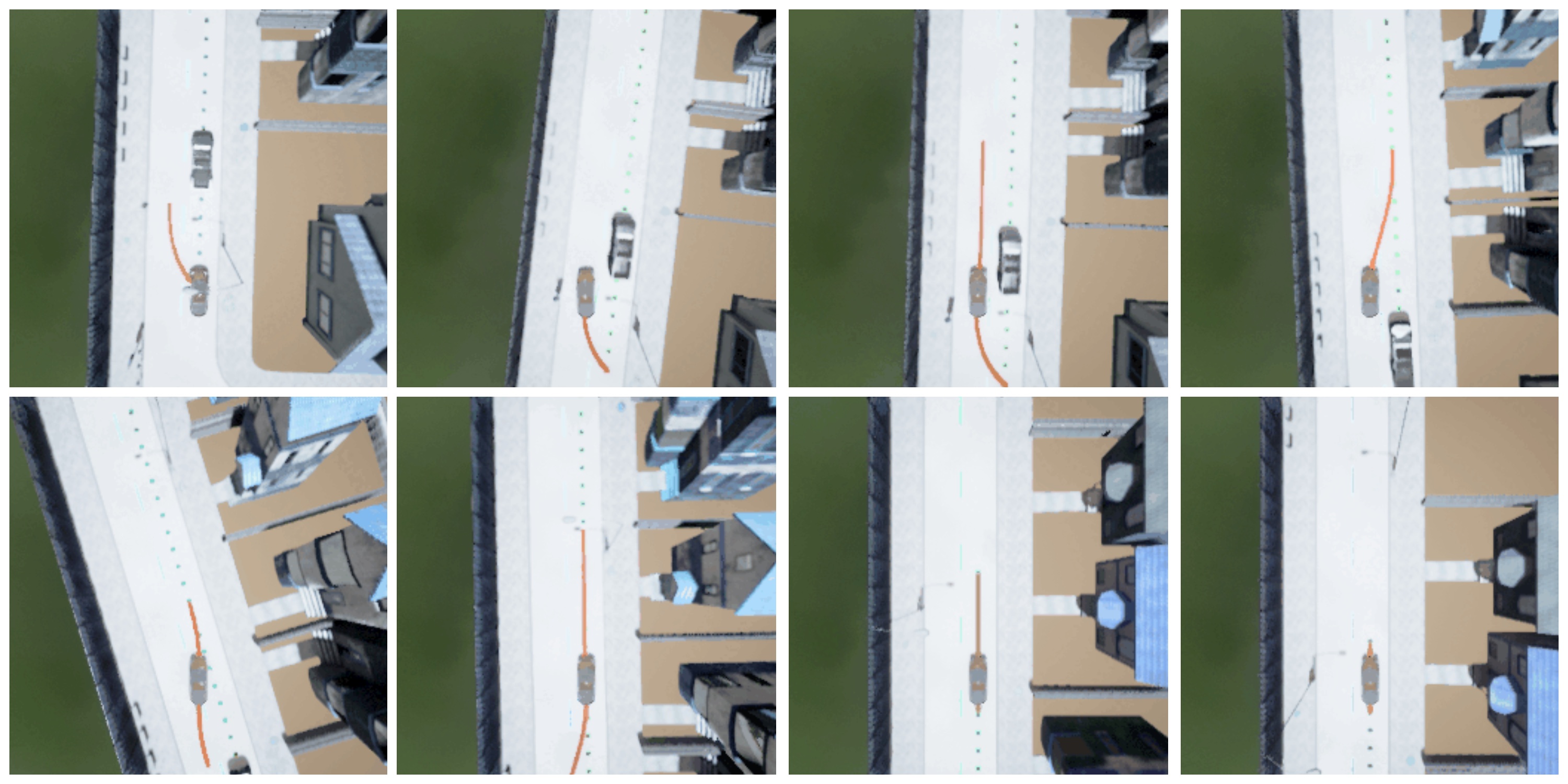}
        \caption{Success case 1}
        \label{fig:s_case_1}
    \end{subfigure}
    \begin{subfigure}[b]{0.49\columnwidth}
        \centering
        \includegraphics[width=\textwidth]{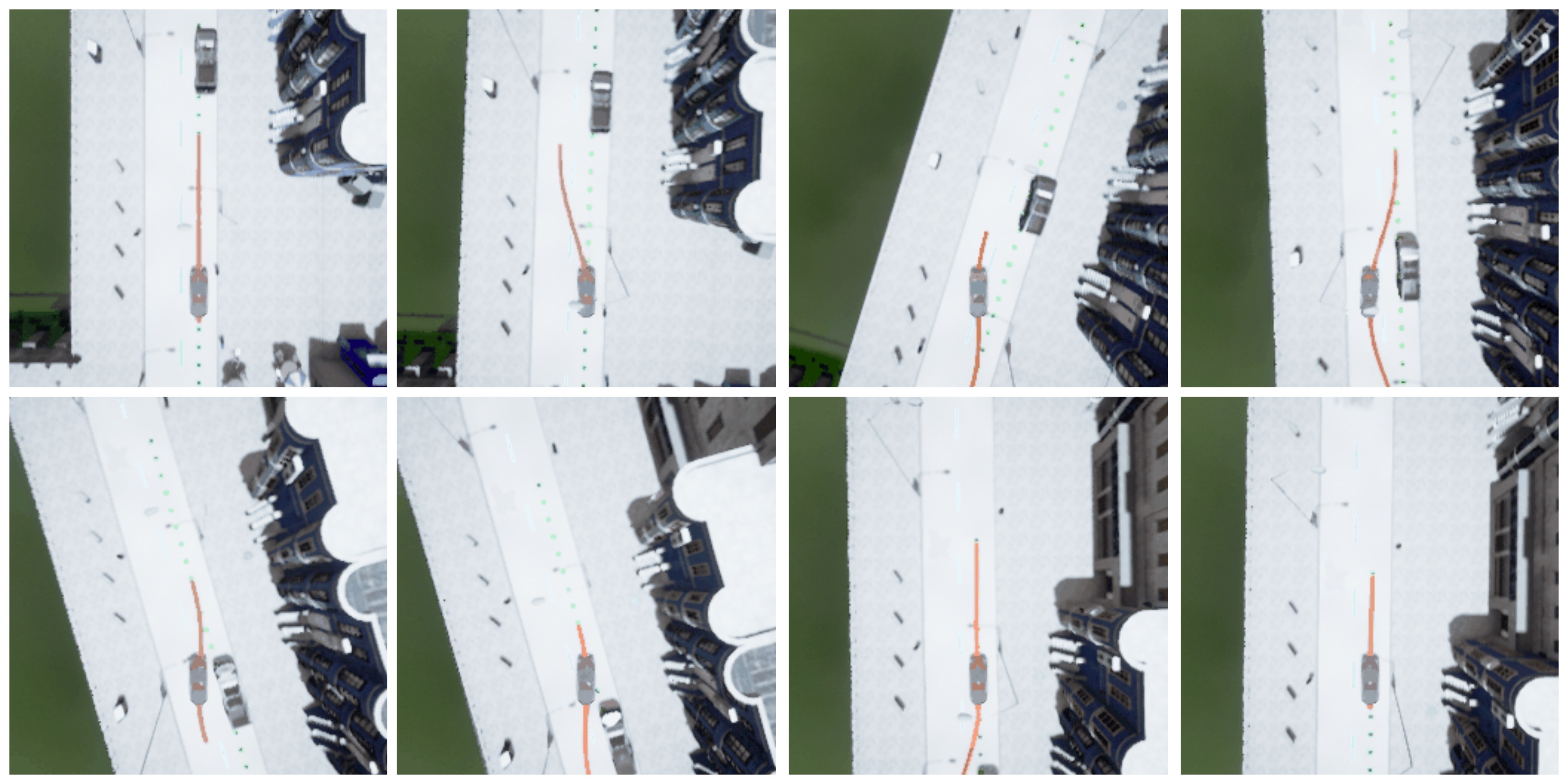}
        \caption{Success case 2}
        \label{fig:s_case_2}
    \end{subfigure}
    \begin{subfigure}[b]{0.49\columnwidth}
        \centering
        \includegraphics[width=\textwidth]{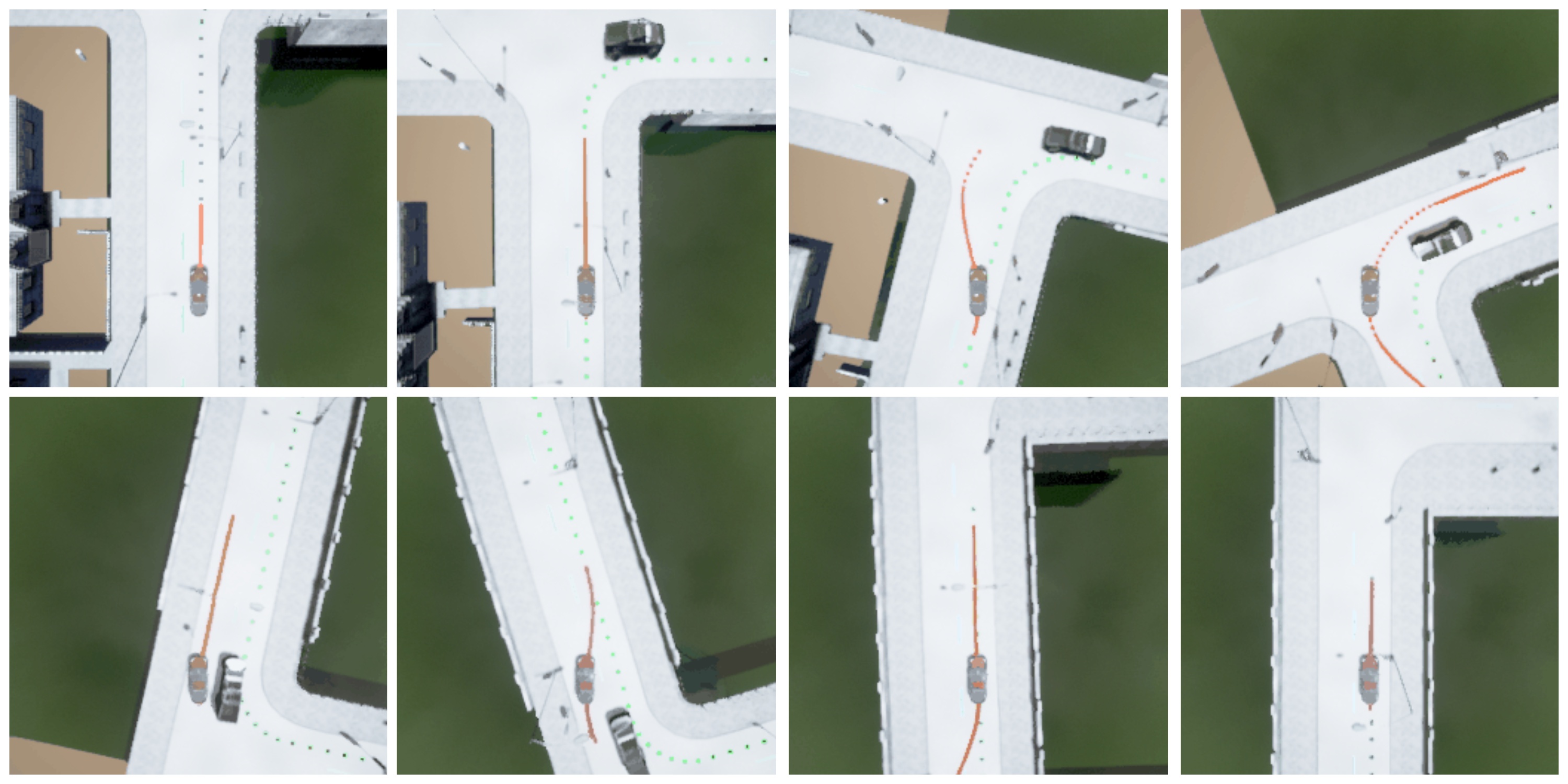}
        \caption{Success case 3}
        \label{fig:s_case_3}
    \end{subfigure}
    \begin{subfigure}[b]{0.49\columnwidth}
        \centering
        \includegraphics[width=\textwidth]{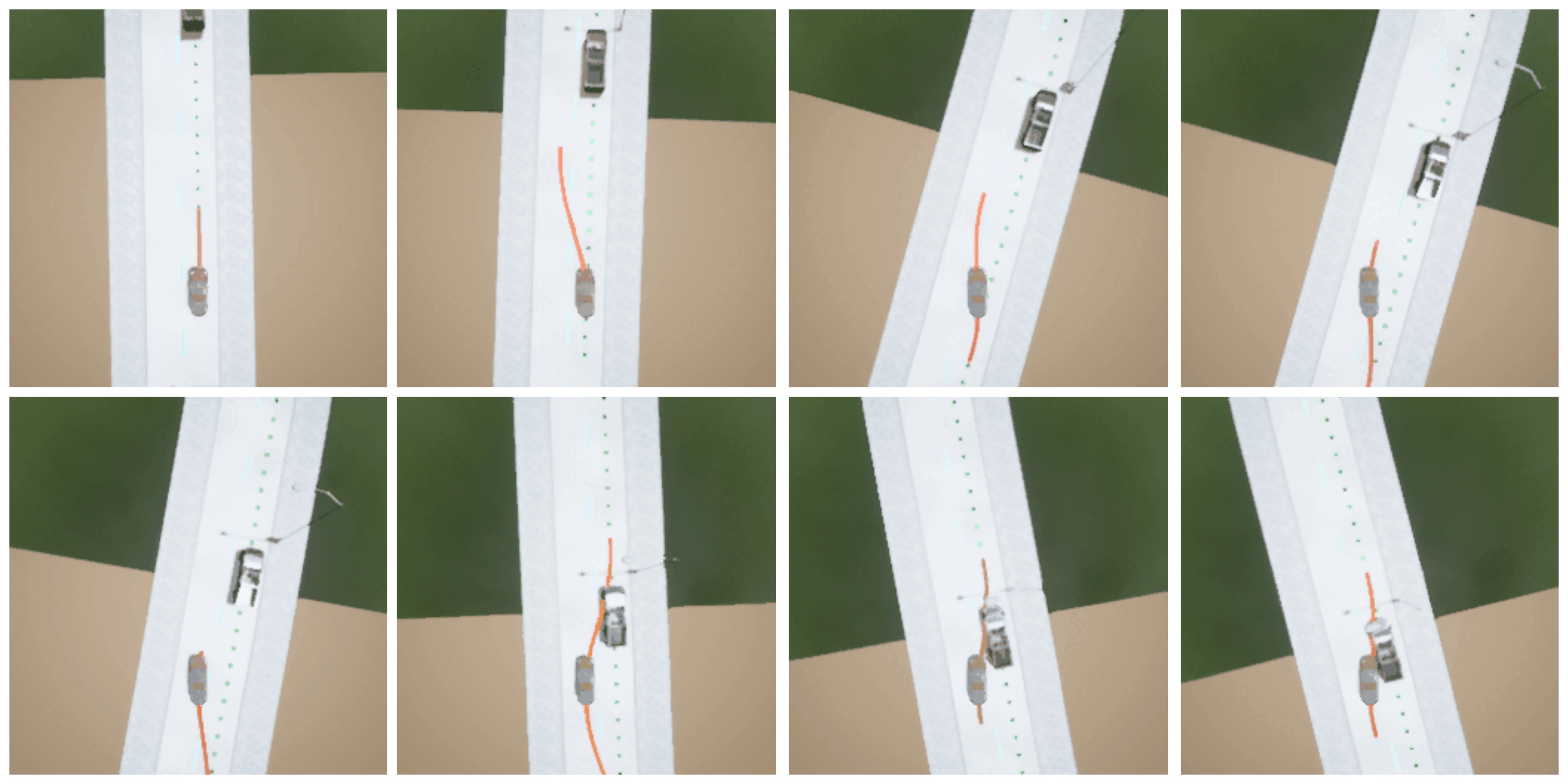}
        \caption{Failure case 1}
        \label{fig:f_case_1}
    \end{subfigure}
    \caption{Our Reinforcement Learning agent has learned to detect situations in which allowed and controlled rule exceptions are necessary in order to progress. Trajectories that avoid obstacles on the road ahead are learned, changing to the oncoming lane temporarily, and returning to the default state as soon as possible. Reprinted from~\cite{Qin_Reinforcement_2023_MA}.}
\label{fig:DreamerV3_prediction}
\end{figure}

\begin{figure}[t!]
    \centering
    \begin{subfigure}[b]{0.49\columnwidth}
        \centering
        \includegraphics[width=\textwidth]{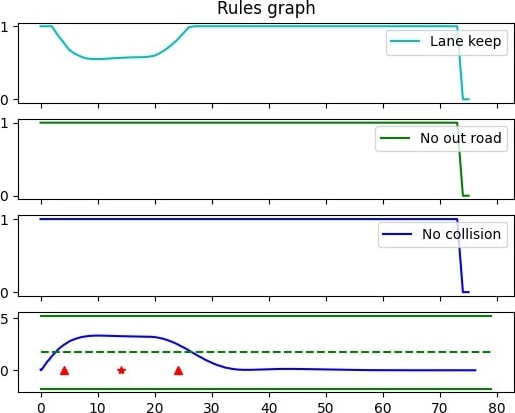}
        \caption{Success case 1}
        \label{fig:sr_case_1}
    \end{subfigure}
    \begin{subfigure}[b]{0.49\columnwidth}
        \centering
        \includegraphics[width=\textwidth]{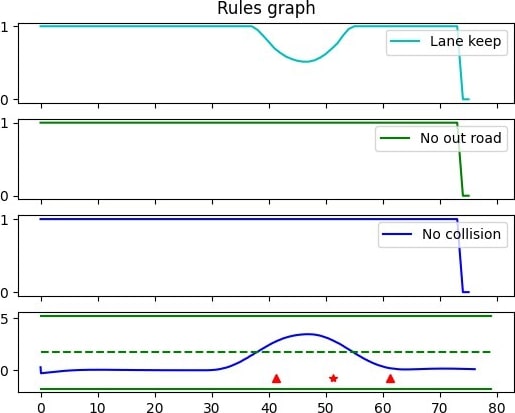}
        \caption{Success case 2}
        \label{fig:sr_case_2}
    \end{subfigure}
    \begin{subfigure}[b]{0.49\columnwidth}
        \centering
        \includegraphics[width=\textwidth]{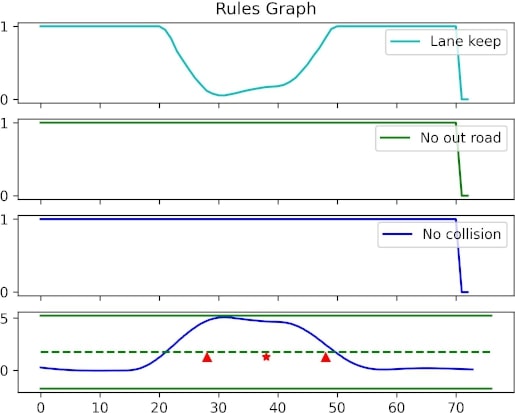}
        \caption{Success case 3}
        \label{fig:sr_case_3}
    \end{subfigure}
    \begin{subfigure}[b]{0.49\columnwidth}
        \centering
        \includegraphics[width=\textwidth]{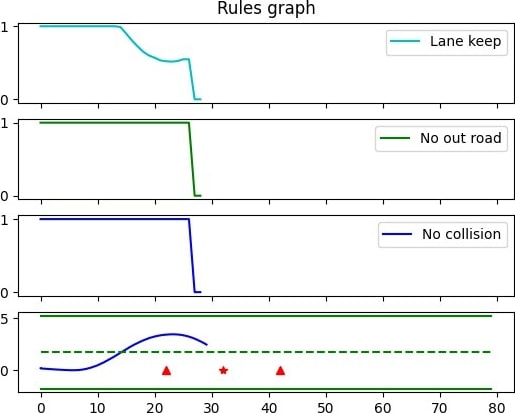}
        \caption{Failure case 1}
        \label{fig:fr_case_1}
    \end{subfigure}
    \caption{Overview of rule adherence for the scenarios shown in Fig.~\ref{fig:DreamerV3_prediction}. The bottom row depicts the scenario, where $\textcolor{red}{\star}$ shows the position of the obstacle and \mytriangle{red} visualize the area in which the rulebook was active. The top three rows show the rule adherence of the analyzed traffic rules, where 1 means full compliance and 0 violation. Reprinted from~\cite{Qin_Reinforcement_2023_MA}.}
    \label{fig:Rule_graph}
\end{figure}

For a better understanding of individual scenarios, we visualize both the agent's driving performance and its adherence to the defined traffic rules as shown in Fig.~\ref{fig:DreamerV3_prediction} and Fig.~\ref{fig:Rule_graph}.

Fig.~\ref{fig:DreamerV3_prediction}, shows raw observations $o_t$ from the BEV camera including the planned trajectory during an episode. This visualization illustrates the vehicle's ability to change lanes to avoid obstacles and then successfully return to the original lane immediately. The corresponding trajectory and traffic rule compliance graphs, plotted in Frenet coordinates, are provided in Fig.~\ref{fig:Rule_graph}. As we have seen from the quantitative results, the agent performs controlled rule exceptions successfully most of the time. However, in most failure cases, we observer too early returns to the original lane of the ego vehicle, as shown in the last scenario.

\section{Conclusion}
\label{sec:conclusion}
In this work, we proposed Informed Reinforcement Learning to perform controlled traffic rule exceptions in autonomous driving. For this, we extend any given RL algorithm by learning trajectories in the Frenet frame, formulating a hierarchical rulebook, and implementing a situation-aware reward design. Our proposed method achieved faster learning convergence than the baseline, displayed robust performance in traffic rule exception scenarios, and successfully incorporated real-world traffic laws into RL reward design via a rulebook. However, there are some limitations to our work. In order to focus on the performance given scenarios that require controlled traffic rule exceptions, we implemented a discrete action space. As already shown in the literature, a continuous action space for the generation of trajectories could be implemented into our work~\cite{moghadamEndtoendDeepReinforcement2020}. We utilize ground-truth information for our situation-awareness, which could be replaced by an independent module. With extensive engineering effort, the number of rules and scenarios can be extended in order to examine the scalability properties of our approach.

However, our results demonstrate that extending an RL agent by both a learned trajectory and a situation-aware reward design accelerates both the training progress and the final performance in challenging scenarios.
\newrobustcmd*{\mytriangle}[1]{\tikz{\filldraw[draw=#1,fill=#1] (0,0) --
        (0.2cm,0) -- (0.1cm,0.2cm);}}

\section{Acknowledgment}
\label{sec:acknowledgment}

This work results partly from the project KI WISSEN (19A20020L), funded by the German Federal Ministry for Economic Affairs and Climate Action (BMWK).

{\small
    \bibliographystyle{abbrv}
    \bibliography{references}
}

\end{document}